\documentclass{article}

\PassOptionsToPackage{numbers,sort&compress}{natbib}

\usepackage[preprint]{neurips_2025}

\usepackage[utf8]{inputenc} 
\usepackage[T1]{fontenc}    
\usepackage{hyperref}       
\usepackage{url}            
\usepackage{booktabs}       
\usepackage{amsfonts}       
\usepackage{nicefrac}       
\usepackage{microtype}      
\usepackage{xcolor}         
\usepackage{graphicx}
\usepackage{amsmath}
\usepackage{subcaption}
\usepackage{placeins}   
\usepackage{amsthm}
\usepackage{amssymb}
\usepackage{xcolor}
\usepackage{cleveref}
\usepackage[font=small]{caption}

\title{Squeezed Diffusion Models}

\author{%
  Jyotirmai Singh \\
  Stanford University\\
  \texttt{joesingh@stanford.edu} \\
  \And
  Samar Khanna \\
  Stanford University \\
  \texttt{samarkhanna@cs.stanford.edu} \\
  \And
  James Burgess \\ 
  Stanford University \\
  \texttt{jmhb@stanford.edu}
}

\begin{document}
\maketitle

\begin{abstract}
Diffusion models typically inject \emph{isotropic} Gaussian noise, disregarding structure in the data. Motivated by the way \emph{quantum squeezed states} redistribute uncertainty according to the Heisenberg uncertainty principle, we introduce \textbf{Squeezed Diffusion Models (SDM)}, which scale noise anisotropically along the principal component of the training distribution. As squeezing enhances the signal-to-noise ratio in physics, we hypothesize that scaling noise in a data-dependent manner can better assist diffusion models in learning important data features. We study two configurations: (i) a \emph{Heisenberg diffusion model} that compensates the scaling on the principal axis  with inverse scaling on orthogonal directions and (ii) a \emph{standard} SDM variant that scales only the principal axis. Counterintuitively, on CIFAR-10/100 and CelebA-64, mild \textit{antisqueezing} -- i.e. increasing variance on the principal axis -- consistently improves FID by up to 15\% and shifts the precision--recall frontier toward higher recall.  Our results demonstrate that simple, data‑aware noise shaping can deliver robust generative gains without architectural changes.

\end{abstract}

\section{Introduction}
Score-based diffusion models have become a standard tool for high-fidelity image generation, achieving state-of-the-art performance
on datasets from CIFAR-10 to ImageNet \citep{ ho2020denoising, dhariwal2021diffusion}. While diffusion models typically rely on adding standard isotropic
Gaussian noise during the training process, there is evidence that 
this standard "one-size-fits-all" approach can degrade generative 
quality in certain contexts like in natural images which have an anisotropic power distribution in frequency space \citep{falck2025fourier}. This has motivated the investigation of methods to engineer the noise used during the diffusion process in a data-dependent way to overcome 
these shortcomings \citep{huang2024bluenoise, sahoo2024mulan}. 

The field of quantum metrology \cite{clerk2008quantumnoise}, has developed an extensive array of techniques to manipulate noise 
in the context of precision measurement. One such technique is quantum
squeezing \citep{walls1983squeezing}, which redistributes noise so that variance is reduced along a “measurement” axis of interest, boosting signal-to-noise ratio (SNR) and enabling weak-signal extraction. Inspired by this, we introduce Squeezed Diffusion Models (SDM), which anisotropically scale noise along the dataset’s principal component. We hypothesize this encourages the model to learn more semantically meaningful patterns during training, improving sample quality in generation.

To test this, we investigated two model variants: (1) a \textit{Heisenberg diffusion model} that scales the principal direction and counter-scales the orthogonal ones and (2) a "standard" SDM
that only scales the principal direction. Surprisingly, we 
observed that best performance came by mildly \textit{antisqueezing} along the principal axis -- i.e. injecting noise and degrading SNR --  across multiple datasets. 

\paragraph{Background} Quantum squeezing exploits the Heisenberg uncertainty principle:
\begin{equation}
    \Delta A \cdot \Delta B \geq \frac{1}{2}\left|\langle\left[A, B\right]\rangle\right|
    \label{heisenberg}
\end{equation} 
by reducing variance in one physical variable $\Delta A$ while enlarging it in the conjugate one $\Delta B$ but keeping the product constant. Squeezed states boost signal-to-noise along a chosen axis and underpin precision measurements such as gravitational-wave  \cite{ligo2024squeezing} and dark matter \cite{haystac2021squeezing} detectors. 

\begin{figure}[t]
    \centering
    \includegraphics[width=1.0\linewidth]{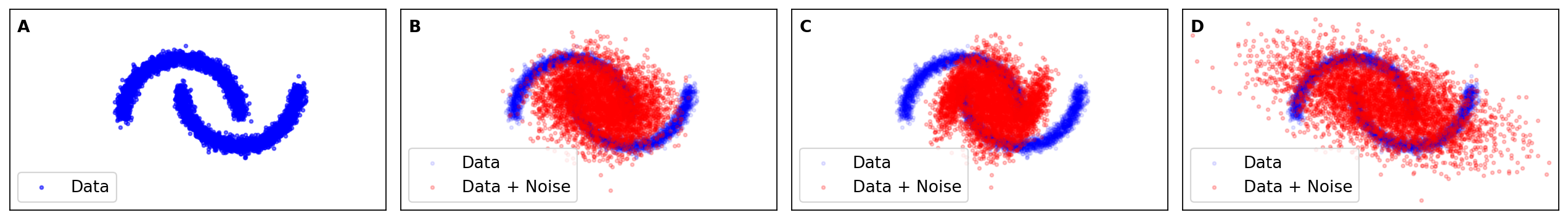}
    \caption{Diffusion with squeezed noise. (a) the raw
    data distribution. (b) standard isotropic noising process. (c)–(d) anisotropic noising process with (c) squeezing and (d) antisqueezing along the principal direction.}
    \label{fig:squeezed_noise}
\end{figure}

\Cref{fig:squeezed_noise} illustrates how this effect can be applied to diffusion models.
In \cref{fig:squeezed_noise}(c) the noise (\textcolor{red}{red}) is squeezed along the first principal component and the original distribution (\textcolor{blue}{blue}) is still somewhat intact due to higher SNR relative to isotropic noise (\cref{fig:squeezed_noise}(b)).
In \cref{fig:squeezed_noise}(d), the noise is instead \textit{antisqueezed} or enhanced along the same direction, reducing SNR compared with the isotropic case.

\paragraph{Related Work}
Recent work has showed that alternative noise schedules can impact generative performance \cite{chen2023importancenoiseschedulingdiffusion}. Improvements have 
been demonstrated with for example the cosine schedule \cite{dhariwal2021diffusion} and variationally optimized schedules \cite{kingma2023variational}. Other work has 
focused on changing noise structure in the frequency domain by e.g. equalizing or
biasing certain frequencies \cite{falck2025fourier, huang2024bluenoise, hao2024wavelet}. Closer to our aim, other work has demonstrated benefits by learning a multivariate data-dependent noise process \cite{sahoo2024mulan}. Our approach differs 
by using a simple, physics-inspired PCA alignment to apply a single-parameter 
anisotropic scaling. 

\section{Method: Squeezed Diffusion}
\vspace{-0.3em}      
\label{sec:method}

As in a standard DDPM, an SDM corrupts a
clean datapoint~$x_0$ with additive Gaussian noise over a schedule
$\{\beta_t\}_{t=1}^{T}\subset(0,1)$. In an SDM however, at each step we \emph{anisotropically} scale the noise with a
\textit{squeeze matrix} $S_t(s)$ whose strength is
controlled by a single hyper‑parameter~$s$ and which may also vary with~$t$:
\begin{align}
  x_t^{\text{sq}}
  &=\sqrt{\alpha_t}\,x_{t-1}^{\text{sq}}
    +\sqrt{1-\alpha_t}\,S_t(s)\,\varepsilon_t,
  &\varepsilon_t &\stackrel{\text{i.i.d.}}{\sim}\mathcal N(0,I),\qquad
    \alpha_t=1-\beta_t .
  \label{eq:forward-step}
\end{align}

\paragraph{Marginal distribution}
By unrolling~\cref{eq:forward-step} we obtain
\begin{equation}
  x_t^{\text{sq}}
    =\sqrt{\bar\alpha_t}\,x_0
     +\sum_{i=1}^{t}
        \Bigl(\sqrt{1-\alpha_i}\,\prod_{j=i+1}^{t}\sqrt{\alpha_j}\Bigr)
        S_i(s)\,\varepsilon_i,
  \qquad
  \bar\alpha_t=\!\!\prod_{i=1}^{t}\alpha_i,
  \label{eq:marginal}
\end{equation}
As in a standard DDPM, we have that $q(x_t | x_0) \sim \mathcal{N}\left( \sqrt{\bar{\alpha}}_t x_0, \Sigma_t\right)$ but instead now the covariance is modified to be
\(
  \Sigma_t
   =\sum_{i=1}^{t}
     \bigl((1-\alpha_i)\prod_{j=i+1}^{t}{\alpha_j}\bigr)
     S_iS_i^{\!\top}.
\)

\paragraph{Training objective}
The network is trained to predict the \emph{squeezed} noise
\(
  \varepsilon_{t}^{\text{sq}}=S_t(s)\varepsilon_t
\)
via the standard MSE:
\begin{equation}
  \mathcal L_{\text{SDM}}
  =\mathbb E_{x_0,t,\varepsilon_t}
    \bigl\lVert
      \varepsilon_{t}^{\text{sq}}
      -\hat\varepsilon_{\theta}^{\text{sq}}(x_t^{\text{sq}},t)
    \bigr\rVert_2^{\,2}.
  \label{eq:training-loss}
\end{equation}

\paragraph{Reverse step (whiten–denoise–resqueeze)}
For the reverse step we first \textit{whiten} each state:

\begin{equation}
  \tilde x_t = S_t^{-1}x_t^{\text{sq}},\qquad
  \tilde\varepsilon_{\theta}
    = S_t^{-1}\hat\varepsilon_{\theta}^{\text{sq}}.
\end{equation}

With this change of variable, \cref{eq:forward-step} becomes \textit{approximately} equal 
to a standard DDPM in \(\tilde x_t\), which allows us to reuse the \citet{ho2020denoising} posterior: 
\begin{equation}
  \tilde x_{t-1}
  =\frac{1}{\sqrt{\alpha_t}}
     \Bigl(
       \tilde x_t
       -\frac{1-\alpha_t}{\sqrt{1-\bar\alpha_t}}\,
        \tilde\varepsilon_{\theta}(\tilde x_t,t)
     \Bigr)
   +\sqrt{\tilde\beta_t}\,z,\qquad
   z\sim\mathcal N(0,I),
   \label{eq:whitened-posterior}
\end{equation}
where
\(
  \tilde\beta_t
  =(1-\bar\alpha_{t-1})/(1-\bar\alpha_t)\,\beta_t
\).
Finally we \textit{re‑squeeze} with the \emph{next} matrix
\(S_{t-1}\) to stay in squeezed coordinates:
\begin{equation}
  x_{t-1}^{\text{sq}} = S_{t-1}(s)\,\tilde x_{t-1}.
\end{equation}
Note that with the whitening, \cref{eq:forward-step} becomes
\begin{equation}
    \tilde{x}_t = \sqrt{\alpha_t}S_t^{-1}S_{t-1} \tilde{x}_{t-1} + \sqrt{1 - \alpha_t}\varepsilon_t 
    \label{eq:whitened-forward-process}
\end{equation}
When $S_t$ is time-independent, equation \cref{eq:whitened-forward-process} is an exact DDPM because $S_t = S_{t-1}$ and \cref{eq:whitened-posterior} is the exact posterior. For time-dependent squeezing this is only an approximate DDPM with some drift $R_t=S_t^{-1}S_{t-1}$. However, for the form of squeezing matrices we define and our linear noise schedule, this drift ends up being negligible so the posterior approximation does not impact generative performance. More details are in  \cref{sec:appendix_b}.

\paragraph{Squeeze matrix $S_t(s)$} To choose the squeezing direction, we do a PCA decomposition of the data in RGB space at the individual pixel level. For large natural image sets, this aligns with the opponent‑colour axes luminance, red–green, and blue–yellow \cite{buchsbaum1983trichromacy, Ruderman:98}. Let $\hat{v}$
be the unit vector in the direction of the first principal component, which for RGB images is the luminance. We
can define the squeeze matrices $S_t^{\mathrm{HDM}}$ for the Heisenberg diffusion model
and $S_t^{\mathrm{SDM}}$ for the standard squeezed diffusion model:
\begin{equation}
\begin{aligned}
  S_t^{\mathrm{HDM}}(s) &= e^{-s}\,\hat v\hat v^{\top}
        + e^{\frac{s}{\,n-1}}\!\bigl(I-\hat v\hat v^{\top}\bigr),
  \qquad
  S_t^{\mathrm{SDM}}(s) &= I+\hat v\hat v^{\top}\bigl(e^{-s}-1\bigr).
\end{aligned}
\end{equation}
Here $s$ is a hyperparameter corresponding to the squeezing
strength and $n$ is the dimension of the vector space on which squeezing acts. For the case of RGB pixels, $n=3$. Note that $S_t^{\mathrm{SDM}}$
is derived from $S_t^{\mathrm{HDM}}$ by just discarding the antisqueezing in the
 subspace orthogonal to $\hat{v}\hat{v}^\top$. These matrices mirror the exponential form of electromagnetic squeezing operators in quantum mechanics \cite{quantumoptics2005}:
 \begin{equation}
     S(z) = \exp\left(\frac{z^*\hat{a}^2 - z(\hat{a}^\dagger)^2}{2}\right)
 \end{equation}
For $z = re^{i\theta}$, $S(z)$ scales the noise by $e^{-r}$ in the $\theta$
direction in the phase space determined by the electromagnetic creation and annihilation operators $\hat{a}^\dagger$, $\hat{a}$, just as the $S_t$ do so by $e^{-s}$ in the direction determined by $\hat{v}$. Note also that with this parametrisation, there is a smooth, continuous path between an SDM and an isotropic DDPM (achieved at $s=0$). This mirrors the way squeezing of vacuum noise is inherently continuous, and it situates SDM as a principled middle ground between isotropic, data-unaware noise and fully data-aware approaches such as covariance-matching diffusion \cite{sahoo2024mulan}.

Finally, to better align the squeezing process
with the noise schedule of the diffusion model, which in our experiments was linear, we made the squeezing 
strength $s$ vary linearly as a function of time $s(t) = s_0\frac{\beta_t}{\beta_{\mathrm{max}}}$ which makes the squeezing matrix $S_t$ time dependent.

\section{Results}
\begin{figure}[htb]
  \centering
  \begin{subfigure}[b]{0.32\textwidth}
    \centering
    \includegraphics[width=\linewidth]{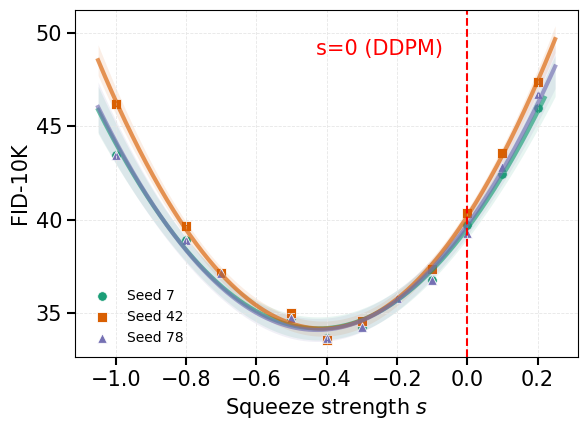}
    \caption{SDM FID vs.\ $s$ on CIFAR-10}
    \label{fig:FIDvsSqueezeCIFAR10}
  \end{subfigure}\hfill
  \begin{subfigure}[b]{0.32\textwidth}
    \centering
    \includegraphics[width=\linewidth]{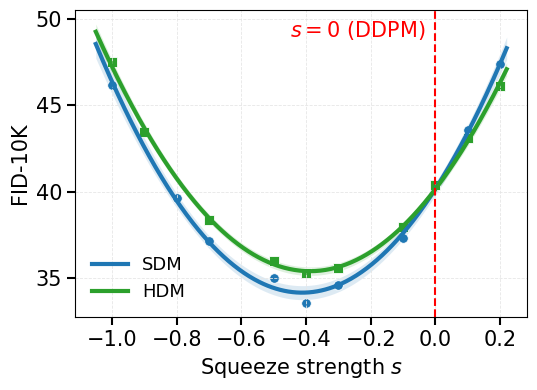}
    \caption{SDM vs.\ HDM FID CIFAR-10}
    \label{fig:FID_SDMvsHDM}
  \end{subfigure}\hfill
  \begin{subfigure}[b]{0.32\textwidth}
    \centering
    \includegraphics[width=\linewidth]{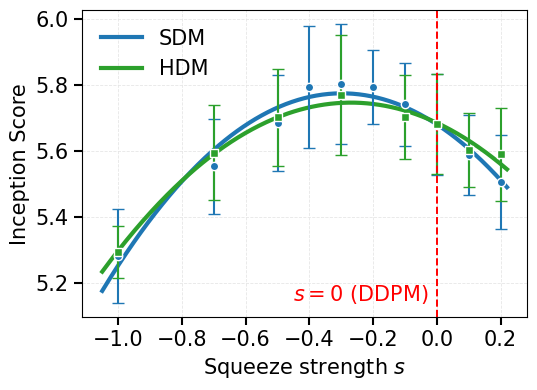}
    \caption{SDM vs.\ HDM IS CIFAR-10}
    \label{fig:is_sdmvshdm}
  \end{subfigure}

  \vspace{2pt}

  \begin{subfigure}[b]{0.32\textwidth}
    \centering
    \includegraphics[width=\linewidth]{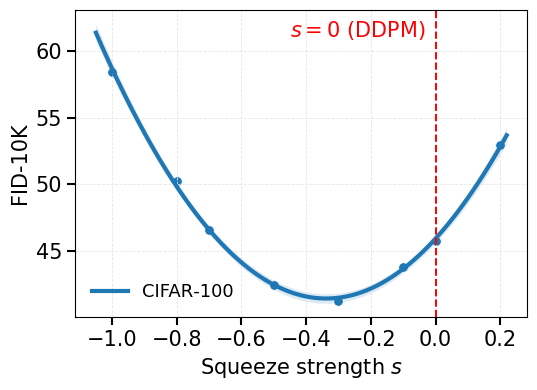}
    \caption{SDM FID vs.\ $s$ on CIFAR-100}
    \label{fig:fid_cifar100}
  \end{subfigure}\hfill
  \begin{subfigure}[b]{0.32\textwidth}
    \centering
    \includegraphics[width=\linewidth]{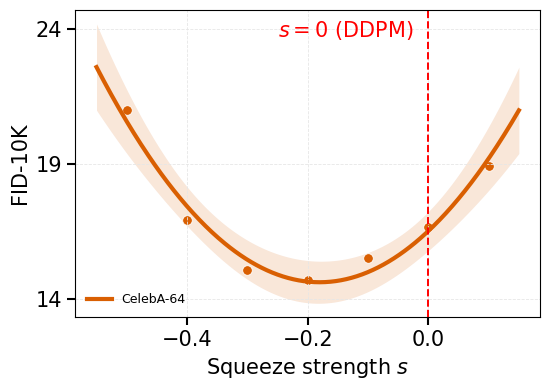}
    \caption{SDM FID vs.\ $s$ on CelebA-64}
    \label{fig:fid_celeba64}
  \end{subfigure}\hfill
  \begin{subfigure}[b]{0.32\textwidth}
    \centering
    \includegraphics[width=\linewidth]{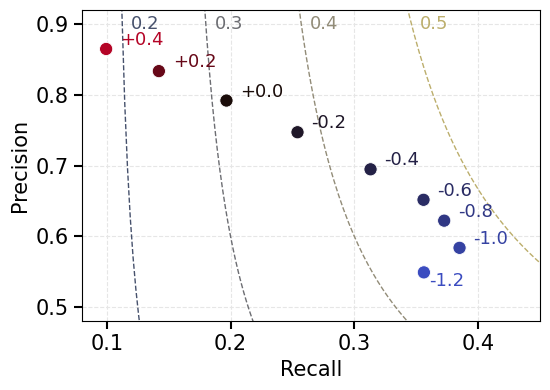}
    \caption{SDM P-R on CIFAR-10}
    \label{fig:pr-plot}
  \end{subfigure}

  \caption{Overview of results across datasets and metrics. (a)–(c) CIFAR-10 FID and IS comparisons for SDM vs.\ HDM; (d)–(e) cross-dataset FID ablations on CIFAR-100 and CelebA-64; (f) precision–recall analysis on CIFAR-10, with squeeze strength
  annotated on points and contours of constant $F$-score.}
  \label{fig:all-results}
\end{figure}

Our experiments consist of ablation studies 
to evaluate squeezed diffusion over a variety of standard image datasets. All experiments were run on a single NVIDIA A100 (40GB) GPU with mixed precision (fp16) via accelerate. The results
are shown in \cref{fig:all-results}.
On CIFAR-10 \cite{krizhevsky2009learning}, FID-10K shows a consistent U-shaped dependence on squeeze strength (\cref{fig:FIDvsSqueezeCIFAR10}) across multiple
random seeds. The average baseline DDPM FID was 39.8 while the SDM reached an FID of 33.6
at $s\approx-0.4$ -- a $\sim$15\% improvement. Notably, this is in the \textit{antisqueezing} regime with degraded
SNR.

Next, with a fixed seed, we compare the standard squeezed diffusion model against the uncertainty preserving Heisenberg variant on CIFAR‑10. Both the FID-10K (\cref{fig:FID_SDMvsHDM}) and Inception Score (\cref{fig:is_sdmvshdm}) follow the same U‑shaped trend in $s$, with optima near $s \approx -0.4$ (FID) and
$ s\approx -0.3$ (IS). SDM outperforms HDM in the mild antisqueezing regime.

To check the robustness of squeezing, we evaluated SDM generative 
performance  on CIFAR-100 \cite{krizhevsky2009learning} in \cref{fig:fid_cifar100} and CelebA-64 \cite{liu2015faceattributes} in \cref{fig:fid_celeba64}. The parabolic shape is reproduced across datasets, with the
optimum achieved in the mild antisqueezing regime again. The
exact location of the optimum varies across datasets, with best performance
on CIFAR-100 at $s \sim -0.3$ and on CelebA-64 at $s \sim -0.2$. 

To gain further insight into the impact of squeezing vs antisqueezing, we analyze the precision and recall metrics of our models on CIFAR-10 \cite{sajjadi2018assessing, kynkaanniemi2019improved, obukhov2020torchfidelity}. \Cref{fig:pr-plot} shows that squeezing the SDM noise corresponds to reducing recall while getting marginal gains in precision. By contrast antisqueezing the noise produces large gains in recall at minor cost in precision up to a point. 
Overall generative quality, as measured by F-score, once again improves up to a point showing parabolic behavior. 

\begin{figure}[!htb]
  \centering
  \begin{subfigure}[b]{0.4\textwidth}
    \centering
    \includegraphics[width=\linewidth]{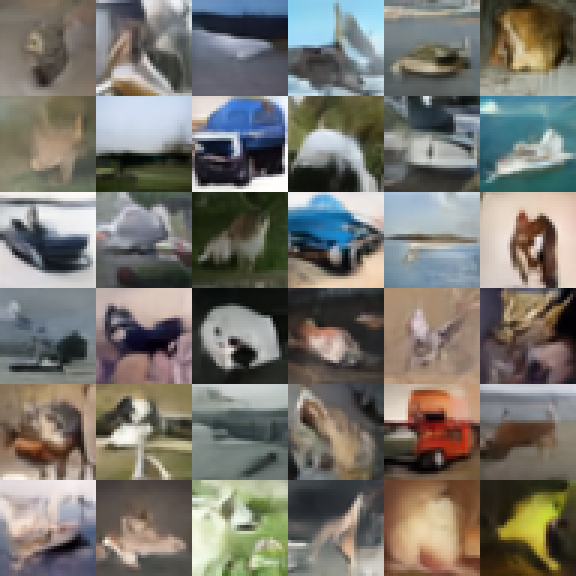}
    \caption{$s=0.0$ (baseline)}
    \label{fig:epoch30_s0}
  \end{subfigure}
  \hspace{5em}%
  \begin{subfigure}[b]{0.4\textwidth}
    \centering
    \includegraphics[width=\linewidth]{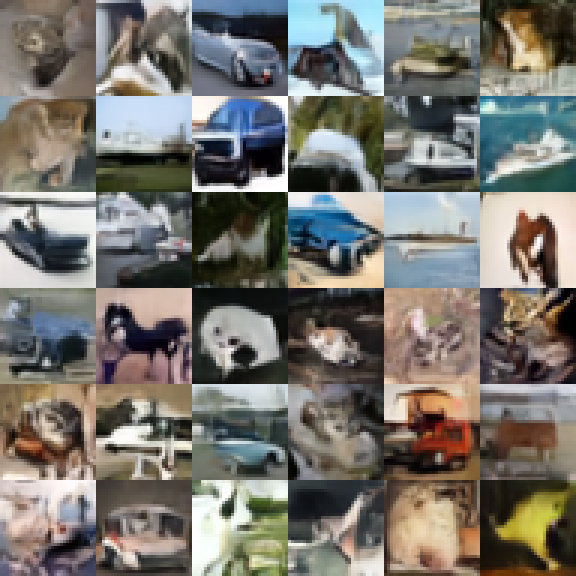}
    \caption{$s=-0.4$ (optimal antisqueezing)}
    \label{fig:epoch30_sm04}
  \end{subfigure}
  \caption{\textbf{Generated CIFAR-10 samples for a standard
  DDPM and an SDM with fixed seed.} (a) standard DDPM; (b) antisqueezed SDM at $s=-0.4$.}
  \label{fig:epoch30_comparison}
\end{figure}

\paragraph{Qualitative Samples} In fig. \ref{fig:epoch30_comparison} we compare 36 samples generated with the same noise seed
for the baseline DDPM and the SDM at the optimal antisqueezing
$s=-0.4$. The SDM images show crisper object contours:
row 1 column 3 becomes a clear car instead of a grey blur,
and row 2 column 6 reveals sharper boat edges and clouds.
These sharper textures are consistent with the
$\sim$15\% FID drop at $s=-0.4$ seen in \ref{fig:FIDvsSqueezeCIFAR10}. Occasional higher frequency ringing artifacts remain
(e.g.\ car in row 2 column 3), echoing the precision loss
seen in the PR analysis as antisqueezing is increased. 

\section{Conclusion and Future Work}
Across three image datasets we observe a stable U‑shaped dependence of sample quality on the squeeze strength: modest antisqueezing enlarges variance along luminance, boosting recall at little precision cost, whereas strong squeezing or antisqueezing harms both.  The standard SDM variant matches or outperforms the Heisenberg counterpart, underscoring that diffusion models need not strictly conserve uncertainty to improve generative performance. Future work could include scaling to higher resolution datasets and exploring frequency-dependent or modality-specific squeezing such as audio. Overall, SDM offers a principled method to manipulate noise that delivers robust gains in generative quality.

\bibliographystyle{unsrtnat}
\bibliography{refs}  

\newpage
\appendix

\section{Details of Model Training}
\label{sec:appendix_a}
\subsection{Implementation}
We use the HuggingFace \texttt{diffusers} library \cite{von_Platen_Diffusers_State-of-the-art_diffusion} with a \texttt{UNet2DModel} backbone. To apply squeezing, we keep the UNet architecture intact and subclass the \texttt{DDPMScheduler} class overriding the \texttt{add\_noise} and \texttt{step} methods. All other training and sampling code remains standard. The model uses a linear noise schedule. Training is done with 1000 timesteps while sampling is done using 50. We also used EMA with a value of EMA decay set to $0.9999$. Code for these experiments is available on Github at \href{https://github.com/joe-singh/squeezing}{\texttt{https://github.com/joe-singh/squeezing}}.

\section{Details of Forward and Reverse Process}
\label{sec:appendix_b}
\subsection{Forward Process}
\begin{proof}
    We prove \cref{eq:marginal} by induction. The statement is true
    for $x_1$ by \cref{eq:forward-step}. Now let us assume it is true for 
    $x_t$. Then:

    \begin{align*}
        x_{t+1} &= \sqrt{\alpha_{t+1}}x_t + \sqrt{1-\alpha_{t+1}}S_{t+1}\varepsilon_{t+1}\\
        &= \sqrt{\alpha_{t+1}}\left[\sqrt{\alpha_t\dots\alpha_1} x_0 + \sum_{i=1}^t S_i \varepsilon_i \sqrt{1-\alpha_i}\prod_{j=i+1}^t\sqrt{\alpha_j}\right]
        + \sqrt{1-\alpha_{t+1}}S_{t+1}\varepsilon_{t+1} \\
        &= \sqrt{\alpha_{t+1}\alpha_t\dots\alpha_1} x_0 + \sum_{i=1}^t S_i \varepsilon_i \sqrt{1-\alpha_i}\prod_{j=i+1}^{t+1}\sqrt{\alpha_j} + \sqrt{1-\alpha_{t+1}}S_{t+1}\varepsilon_{t+1} \\ 
        &= \sqrt{\bar\alpha_{t+1}} x_0 + \sum_{i=1}^{t+1} S_i \varepsilon_i \sqrt{1-\alpha_i}\prod_{j=i+1}^{t+1}\sqrt{\alpha_j}
    \end{align*}
\end{proof}

\subsection{Validity of Posterior Approximation}
In the text we noted that for time dependent squeezing, the Ho \textit{et al.} 
posterior is an approximation. To see why, we compute the drift factor $S_t^{-1}S_{t-1}$ for the standard SDM case. We first compute the inverse of $S_t^{\mathrm{SDM}}$ using the Sherman-Morrison
formula:

\begin{equation}
    S_t^{-1} = I + vv^\top\left(e^{s(t)} -1\right) \label{eq:sdm_inverse}
\end{equation}

In our work $s(t)$ follows a linear schedule given by $s(t) = s_0 \frac{\beta_t}{\beta_{\mathrm{max}}}$. Computing $S_t^{-1}S_{t-1}$ with this linear schedule yields

\begin{equation}
    S_t^{-1}S_{t-1} = I + vv^\top\left(\exp\left[\frac{s_0\Delta\beta}{\beta_{\mathrm{max}}}\right] - 1\right)
\end{equation}

Where $\Delta\beta = \beta_t - \beta_{t-1}$. The \texttt{diffusers} default linear schedule begins at $\beta = 1\times10^{-4}$ and
ends at $\beta = 0.02$. With 1000 timesteps (training), $\Delta\beta \sim 2\times 10^{-5}$ 
while with 50 timesteps (inference) $\Delta\beta \sim 4\times 10^{-4}$. Both numbers are small enough such that for small squeezing strengths $|s_0|$ the exponential
is close to 1 so the drift term is approximately the identity and the posterior
approximation in equation \cref{eq:whitened-posterior} holds.

\end{document}